**Title**: Deep learning prediction of patient response time course from early data via neural-pharmacokinetic/pharmacodynamic modeling

**Authors:** James Lu[1*], Brendan Bender[1], Jin Y. Jin[1], Yuanfang Guan[2]

**Affiliations**
[1] Department of Clinical Pharmacology, Genentech, Inc., South San Francisco, California 94080, USA
[2] Department of Computational Medicine and Bioinformatics, University of Michigan, Ann Arbor, Michigan 48109, USA
[*]Corresponding author, lu.james@gene.com

**Abstract**

The longitudinal analysis of patient response time course following doses of therapeutics is currently performed using Pharmacokinetic/Pharmacodynamic (PK/PD) methodologies, which requires significant human experience and expertise in the modeling of dynamical systems. By utilizing recent advancements in deep learning, we show that the governing differential equations can be learnt directly from longitudinal patient data. In particular, we propose a novel neural-PK/PD framework that combines key pharmacological principles with neural ordinary differential equations. We applied it to an analysis of drug concentration and platelet response from a clinical dataset consisting of over 600 patients. We show that the neural-PK/PD model improves upon a state-of-the-art model with respect to metrics for temporal prediction. Furthermore, by incorporating key PK/PD concepts into its architecture, the model can generalize and enable the simulations of patient responses to untested dosing regimens. These results demonstrate the potential of neural-PK/PD for automated predictive analytics of patient response time course.

**Introduction**

The longitudinal analysis of patient response time course following doses of therapeutics is an important topic for drug development and personalized medicine. However, this is a highly challenging task due to the variability of patient response to treatments and the complexity of drug mechanisms. In response to the need for better characterizing the dynamics between dosing, drug concentration and patient response, the discipline of Pharmacokinetic/Pharmacodynamic (PK/PD) has been developed to link systemic drug concentration kinetics (i.e., PK) to the resulting pharmacological effects over time (i.e., PD) using mathematical models [1]. These models enable the description and prediction of the time course of physiological effects (e.g. tumor size, platelet count, etc.) in response to drugs of various dosage regimens [1, 2]. PK/PD models are typically developed using ordinary differential equations (ODEs), the construction of which have relied upon human modelers' abstraction of data into dynamical systems. Currently, the state-of-the-art methodology for estimating the set of model parameters in PK/PD models is the population approach [3], whereby certain statistical distributions of parameters and error models are assumed and subsequently iterative optimization techniques are applied to computationally minimize the discrepancy between observed and predicted trajectory in some appropriate error metric [4, 5, 2]. The performance of alternative models is compared, and a selection is ultimately made based on various diagnostic criteria and modeling statistics [4, 5]. As described above, the population-PK/PD (pop-PK/PD) modeling paradigm is highly iterative in nature. The accuracy of such models for making temporal predictions depends on the modeler's ability and expertise in describing complex data sets with mathematical equations. However, as the range of data modalities increase in modern biomedical applications to include imaging, high dimensional assays, and continuous monitoring devises, it becomes ever more challenging for the human modelers to glean insight from such large volumes of data. Coupled to the growing data is the need for improving PK/PD models' ability to perform temporal extrapolations, which is key to precision dosing applications [6]. Motivated by these challenges, we explore the possibility of using deep learning to build PK/PD models to augment human capabilities in abstracting dynamical systems from patient data following drug treatment while enhancing predictive accuracy.

With the recent development of the neural ordinary differential equations (neural-ODE) methodology [7], it has become possible to consider an ODE modeling paradigm whereby one can learn the governing equations algorithmically and directly from the data [8, 9]. In particular, this novel machine learning approach generates the input-to-output mapping as the numerical integration of an ODE system described by a neural network, in which the backpropagation is carried out by an adjoint solution method [7]. The neural-ODE methodology is well suited for time-series analysis and especially in the PK/PD setting, as both the dosing and measurement times can be irregular. In particular, it has been applied to fields such as the life sciences [10], image processing [11] and computational physics [12, 13, 14]. While there are several examples of neural-ODE methodologies that have been developed and applied to publicly available biomedical data sets (such as PhysioNet) [10, 15, 16], there is currently no implementation in which the dosing of a therapeutic drug and its concentration data are both explicitly represented in the model. In addition, as these model formulations do not place constraints on the form of ODE systems being constructed and lack the incorporation of key pharmacological principles into their architecture, their ability to generalize from the training set and predict unseen dosing regimens remain in question.

In this work, we propose a new deep learning approach to build PK/PD models that directly learn the governing equations from data with the aim of predicting patient response time course as well as being able to simulate the effects of unseen dosing regimens. Our approach is novel in two ways. Firstly, the architecture of our deep learning model ensures that the pharmacological principle of *dose-concentration-effect* is preserved; that is, the model assumes the causal relationship of dosing driving the drug concentration, which in turn drives the effect dynamics. Secondly, the network architecture is constructed in a manner such that the dosing data enters the model via not one but two ports, thereby ensuring the ability of the model to not only predict the existing treatment data but also enable the simulation of "what-if" scenarios whereby the patients' dosing regimen is modified. Our work illustrates how by incorporating key domain specific modeling principles into the neural network architecture, human and machine intelligence can work together in building dynamical systems that are more likely to generalize well beyond the training data and be better embraced by the domain experts.

Herein, we detail a sequential methodology for the construction of a neural-PK model and a neural-PK/PD model. We demonstrate the approach in describing and predicting drug concentration and platelet dynamics following trastuzumab emtansine (T-DM1) treatment, which is an approved anti-cancer therapy (intravenous administration at 3.6 mg/kg once every three weeks [Q3W]) for the treatment of human epidermal growth factor receptor 2 (HER2)–positive metastatic breast cancer in patients failing prior treatment with trastuzumab and a taxane [17]. Thrombocytopenia, a decrease in platelets requiring dose reductions and delays, is the dose limiting toxicity for T-DM1 [17]. To show the utility of this methodology, we benchmark against the original pop-PK/PD analyses [18, 17], which utilized the myelosuppression PKPD modeling approach proposed by Friberg *et al* [19] that is considered the "gold-standard" for modeling such clinical data. We develop a neural network to search within a space of ODE systems no larger in dimension than that of the existing pop-PK/PD model [17] and compare the resulting neural-PK/PD model with pop-PK/PD in terms of their ability to predict future platelet counts from early data at the individual patient level. Finally, we illustrate the generalizability of neural-PK/PD predictions by performing simulations to predict the effects of alternate (and untested) dosing regimens.

**Results**

*Overview of Model*

We incorporate the basic principles of PK/PD [1] into our neural-PK/PD architecture (Fig. 1) by encoding the following standard PK/PD assumptions: (1) the dynamics of PK is driven by dosing and is independent of PD; (2) the dynamics of PD is driven by PK as well as by itself [1]. We built the above assumptions into the computational graph of the model via the explicit representation of the "Dose" input to the ODE sub-module as shown in Fig. 1(a), and in the dependence of the PK and PD vector fields shown in Fig. 1(c) and (d). In particular, the "Dose" data enters as Dirac delta forcing to the PK component of the ODE system (refer to the Methods section and Supplementary Fig. 1(b) for further details). This precise treatment of dosing is a key part of our model implementation that follows the traditional PK/PD modeling paradigm and which helps to ensure the generalizability of the model predictions to unseen dose schedules. The PK components of the ODE vector field are not influenced by the PD states, whereas the PD components of the ODE vector field are influenced by both the PK and PD states. The individual patient data (including the observed drug concentration, dosed amounts and pharmacodynamic response) is fed into the network port "PKPDData"; see the Methods section for more details. An innovative feature of our model is that drug dosing enters the network at two locations: as part of the "PKPDData" and also through the "Dose" port. While the dosing data that forms part of "PKPDData" represents what doses patients have been treated with, the dosing that enters "Dose" port represents what the model uses in making predictions. For the purpose of reproducing the data in the training set, the doses that enter the two separate ports are identical. However, once the model has been trained and a novel dosing regimen is to be tested, the dosing data that enters the "Dose" port is modified as desired. For details on the training and the structure of the neural-PK model, please refer to the Methods section.

## *PK (drug concentration) data is recapitulated almost perfectly*

In the current work, the PK and PD parts of the network are built in a sequential manner [20], as is the case for the original pop-PK/PD model [18, 17] that we benchmark against. In particular, we take a pop-PK model that was built previously for T-DM1 [21, 22] and train a neural-PK model to mimic the pop-PK model by reading in the early PK data and predict the future time points. We used T-DM1 treatment dataset involving 665 patients, with a median observation and dosing record of 169 days (minimum: 1 day, maximum: 862 days). We split the total number of available patients into a training and testing set. For details on the training and the structure of the neural-PK model, please refer to the Methods section. As illustrated in Fig. 2, the trained neural-PK model can effectively predict (on unseen test patients) the complete time-course of the pop-PK model for T-DM1 drug concentration, by using only the observed PK data up to day 21, which is the first cycle of the treatment. For a quantitative assessment of the neural-PK model, we evaluated it on the n=133 test patients with 3228 PK observations made from times t ≥ 21 days. The predicted drug concentrations from the neural-PK model are in a good correspondence with the predicted drug concentrations from the original pop-PK model, with r-squared ($r2$) = 0.98, correlation coefficient = 0.99 and the root-mean-squared-error (RMSE) of 2.67.

*Qualitative dynamics of PD (drug response) is recapitulated*

By using the trained neural-PK sub-module to drive the neural-PD sub-module, we train the neural network to generate a dynamical system that recapitulates the observed patient platelet trajectories (see the Methods section for details). Fig. 3 shows a comparison of predictions from the neural-PK/PD model to the pop-PK/PD model [17] with the actual platelet data for randomly selected example patients from the test set. In particular, the observed drug concentration and platelet data from the first dosing cycle (i.e., up to 21 days) are provided to the neural network, which subsequently generates the predicted drug concentration and platelet response for all future time points (after 21 days) based on the individual dosing information. The result shows that by incorporating PK/PD principles into its network architecture but otherwise without additional human input, the proposed methodology is able to generate a PK/PD model that demonstrates comparable qualitative dynamical behavior to the state-of-the-art model [17]. In particular, as shown in Fig. 3 the neural-PK/PD model predictions demonstrate a drop in the platelet count post dosing, the subsequent recovery dynamics, as well as a gradual decrease in both the peaks and nadirs of platelet counts in some patients who have been treated for long durations. While significant human expertise and effort was required to build a complex PK/PD model [17] (containing negative feedback, time-dependent and nonlinear terms) based on the Friberg formalism [19], a qualitatively similar dynamical system was constructed in an automated fashion by applying neural-PK/PD modeling directly to data.

*Neural-PK/PD model outperforms the current "gold-standard"*

We benchmark our neural-PK/PD model against the current "gold-standard" pop-PK/PD model [17]. We considered the following 3 sets of observation limits: tObs = 21, 42 and 63 days. In each case, we use the data within the initial observation window (t < tObs) in order to predict all the future platelet data (t $\geq$ tObs). Fig. 4 presents a comparison of pop-PK/PD and neural-PK/PD model predictions versus the observed data ("ground truth"), demonstrating that the latter surpasses the former using both the r2 and RMSE measures of prediction performance. In fact, as shown in Table 1 (Cases (a) to (c)), for all observation windows (Obs = 21, 42 and 63 days) the performance of neural-PK/PD surpasses that of pop-PK/PD by a sizeable margin.

What are the potential implications of improving the prediction accuracy using neural-PK/PD modeling as compared to the current pop-PK/PD modeling approaches? We considered the following scenario: suppose we wish to predict the individual platelet counts in test patients, over the time horizon of t $\geq$ 42 days. Using the existing pop-PK/PD approach whereby data is observed for tObs < 42 days, we obtain r2 = 0.39 and RMSE = 59.8 (refer to Case (b) in Table 1). In contrast, using the proposed neural-PK/PD approach we can make more precise predictions (r2 = 0.45, RMSE = 56.36) using only half the observation window, tObs < 21 days (refer to Case (d) in Table 1). The ability of the neural-PK/PD model to use less data while improving the precision of forecast for patients' future response as compared to the current state-of-the-art model makes it a promising methodology that could potentially enable more robust predictive analytics based on earlier observation. To our knowledge, this is the first demonstration of a deep learning model that surpasses the state-of-the-art pop-PK/PD model in terms of predictive performance.

*Simulating alternative dosing regimen*

One key use of pop-PK/PD model is to perform simulations of new dosing regimens of interest [4, 5] and predict the corresponding drug concentration and response of such untested scenarios in patients. By design, dosing enters neural-PK/PD model as an independent input thereby ensuring that different dosing schedules can be evaluated as desired. As a demonstration of the utility of neural-PK/PD models to predict alternative dosing regimens, we show in Fig. 5 the model simulations for once every 3 weeks (Q3W) dosing at 3.6 mg/kg of T-DM1, once every week (Q1W) dosing at 2.4 mg/kg of T-DM1, as well as once every 3 days (Q3D) dosing at 1.0 mg/kg of T-DM1, for the patients in the training set. Note that the majority of the patients in the training set were treated with Q3W dosing, with the exception of a handful of patients who were treated with Q1W dosing. In particular, no patient was given the Q3D dosing hence it is a dose schedule the model has never been trained on. The dosage of 1.0 mg/kg for the Q3D schedule was chosen to be equivalent to the 2.4 mg/kg Q1W based on the weekly total dose (i.e., $2.4 \times \frac{3}{7} = 1.029$). The simulation results shown in in Fig. 5 demonstrates a larger peak-to-nadir swing of platelet counts in the Q3W dosing as compared to Q1W dosing but with comparable nadir values, a finding that is consistent with the pop-PK/PD simulation results [17]. Furthermore, the PK accumulation observed for the Q3D dosing is consistent with the expected PK behavior of T-DM1. The ability of the model to generate simulations with previously unseen dosing stems from the neural network architecture having been designed based on PK/PD principles, resulting in meaningful extrapolations.

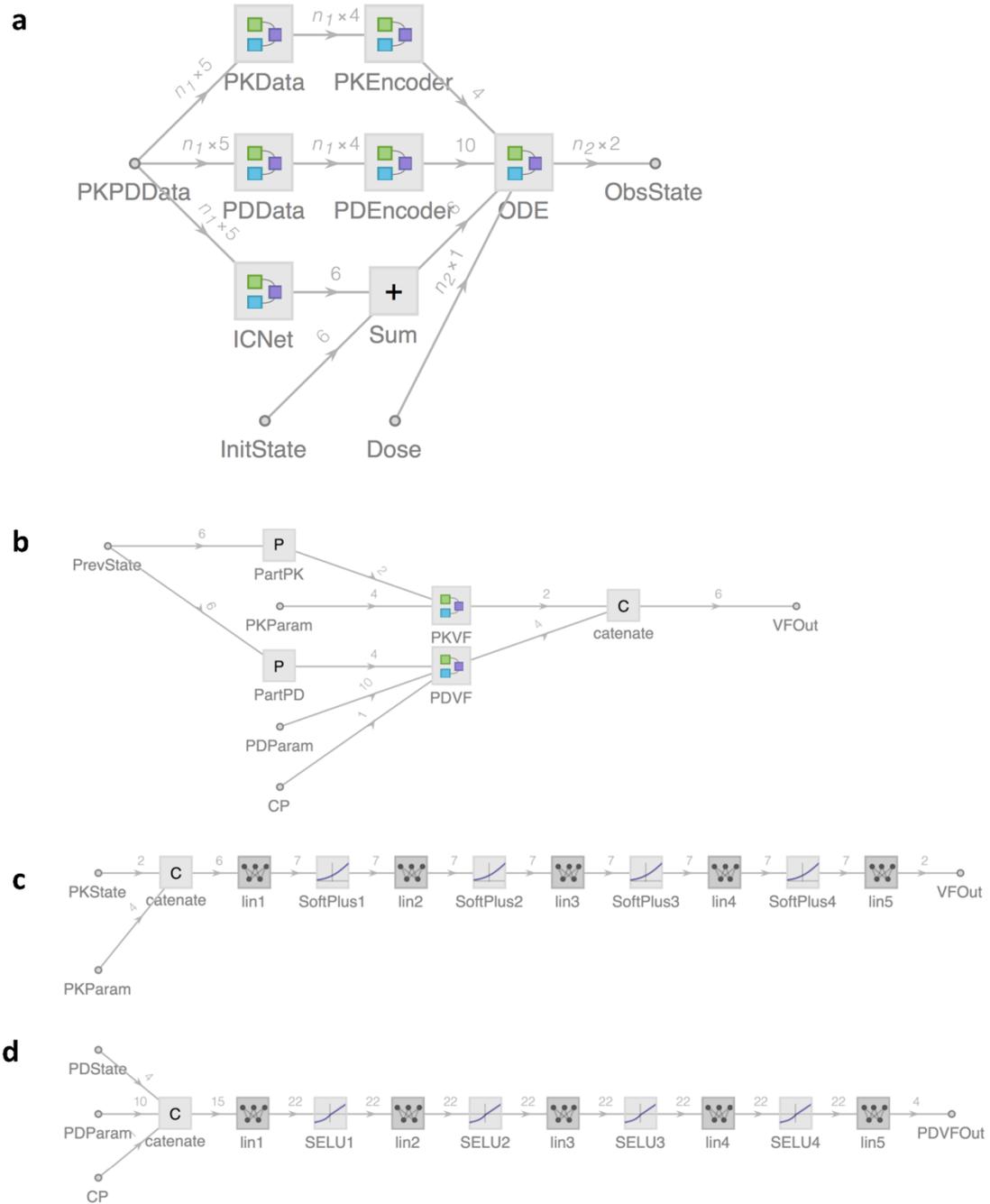

Figure 1. Schematic diagram of the neural-PK/PD model and the vector field (VF) sub-modules. (a) The input data (PK, PD and dosing) is passed as a table of numeric values into the network port "PKPDData", which is subsequently encoded into vectors of various dimensions along 3 parallel pathways: (1) "PKData" → "PKEncoder" feeds into the PK sub-module and determines the PK time course; (2) "PDData" → "PDEncoder" feeds into the PD sub-module and determines how the PD variables change with time in accordance with PK; (3) "ICNet"→ "Sum" feeds in the initial condition of the ODE system. The "ODE" sub-module is a recurrent neural net that unfolds along the time dimension to produce a sequence of predictions for PK and PD. The model generated PK and PD predictions are sent to the network port "ObsState". (b) The inputs to the VF network consist of the current value of the state vector (denoted as "PrevState") as well the PK and PD parameters from the respective encoder networks. The computation of "PKVF" is performed using the PK components of the state vector and the PK parameters. The computation of "PDVF" is performed using the PD components of the state vector, the PD parameters and the drug concentration in patients' plasma "CP". The results are subsequently concatenated and sent to "VFOut". (c) "PKVF" consists of linear and "SoftPlus" [23] layers. (d) "PDVF" consists of linear and "SELU" [24] layers. The numbers in grey denote the dimensions of the tensors/vectors that are being sent.

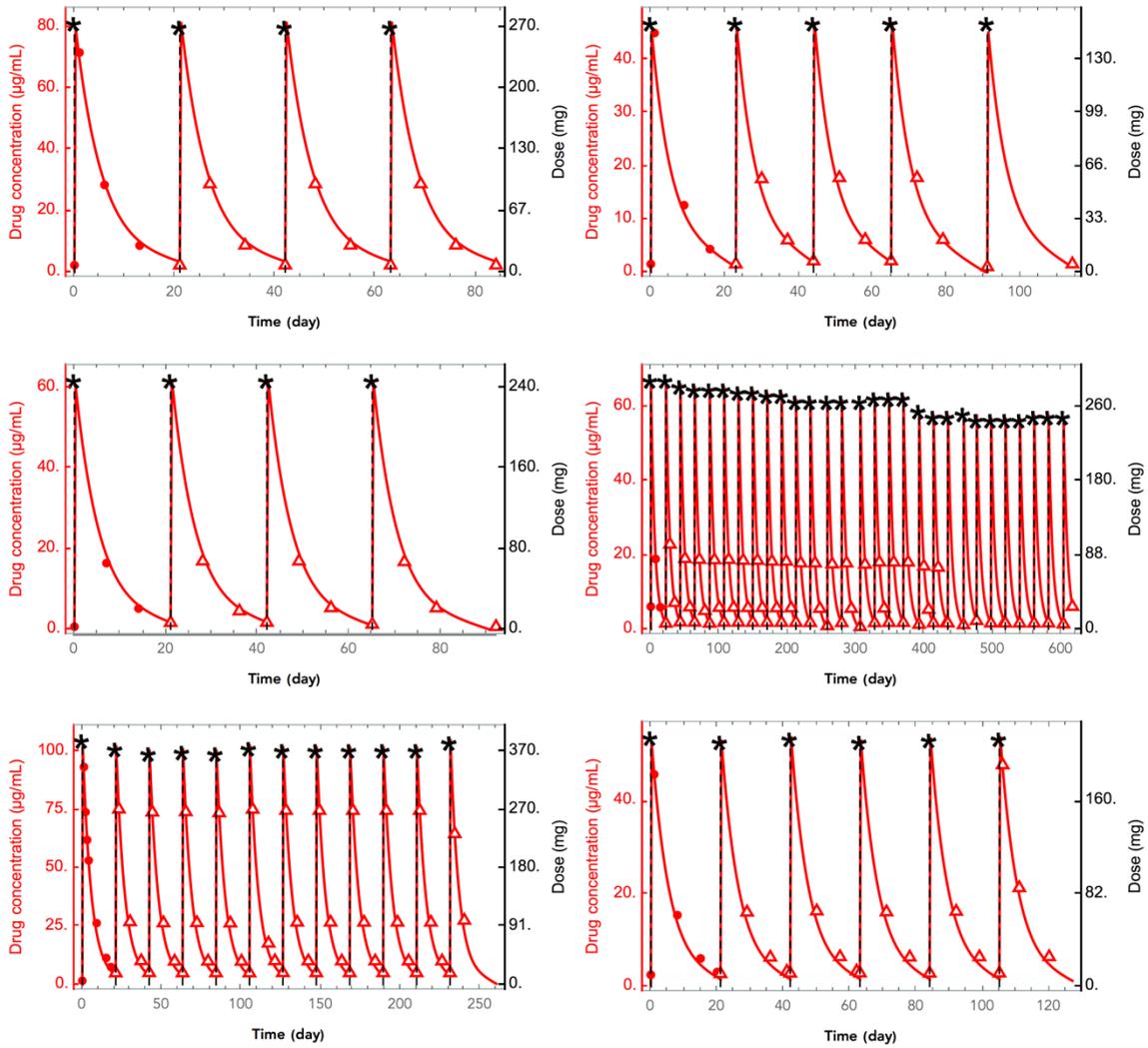

*Figure 2. Illustration (via randomly selected 6 test patients) of the ability of neural-PK model for predicting complete PK profiles from having being shown only the early PK (drug concentration) data from the first dosing cycle (time < 21 days). For the case shown here, PK data within the observation time window t ∈[0, 21] is passed onto the PK encoder network, and the model predicts the PK time course given the complete dosing data. The black stars are the dosages, the filled red circles are the observed drug concentration data for t < 21, the open red triangles are unobserved drug concentrations for t ≥21 days and the neural-PK model predictions are shown as the solid red curves.*

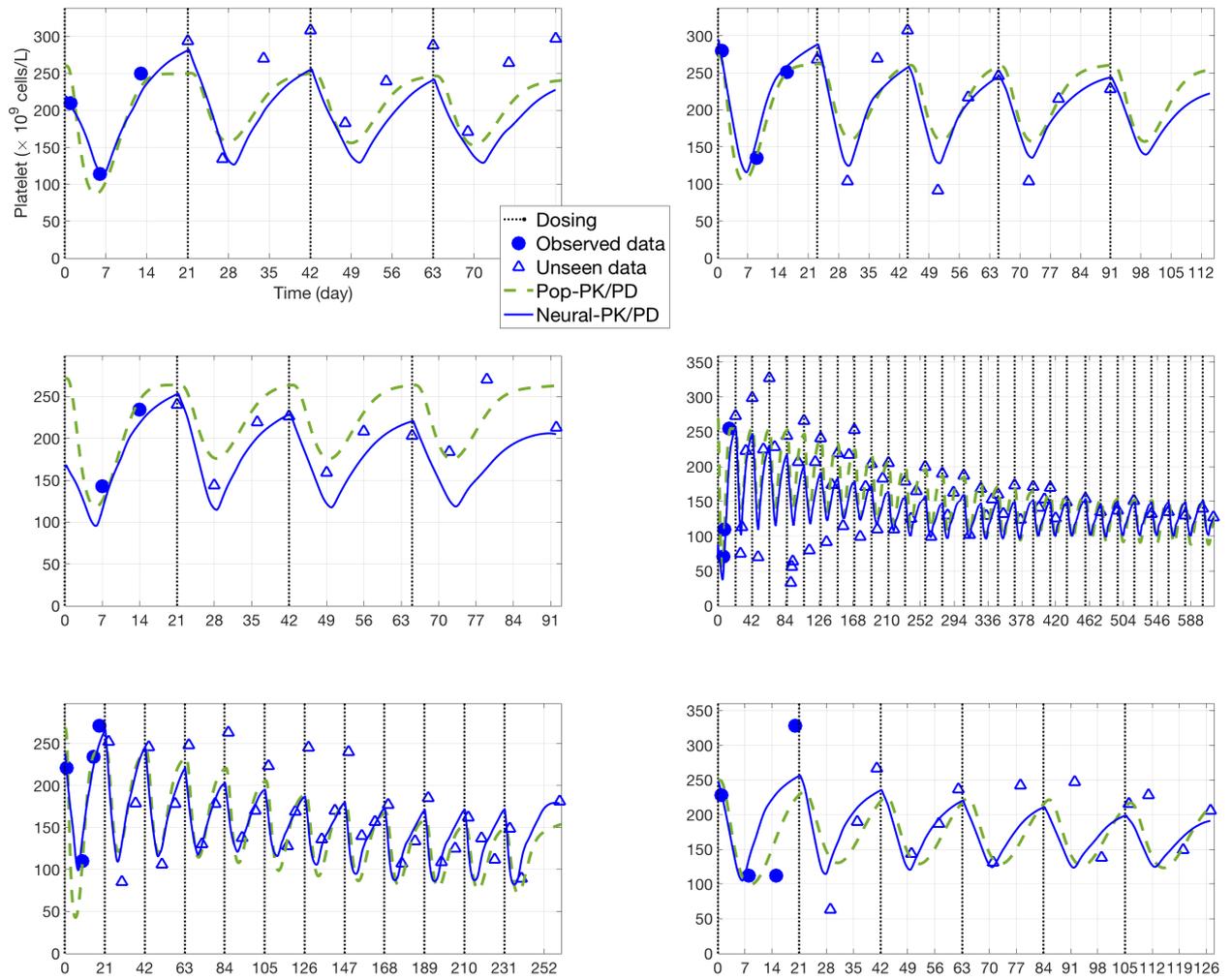

*Figure 3.* Comparisons of platelet predictions from pop-PK/PD model versus neural-PK/PD model in selected patients from the test set, whereby only data from t < 21 days is observed. The plots show similar qualitative dynamics between the two models, as well as some quantitative differences in their predictions.

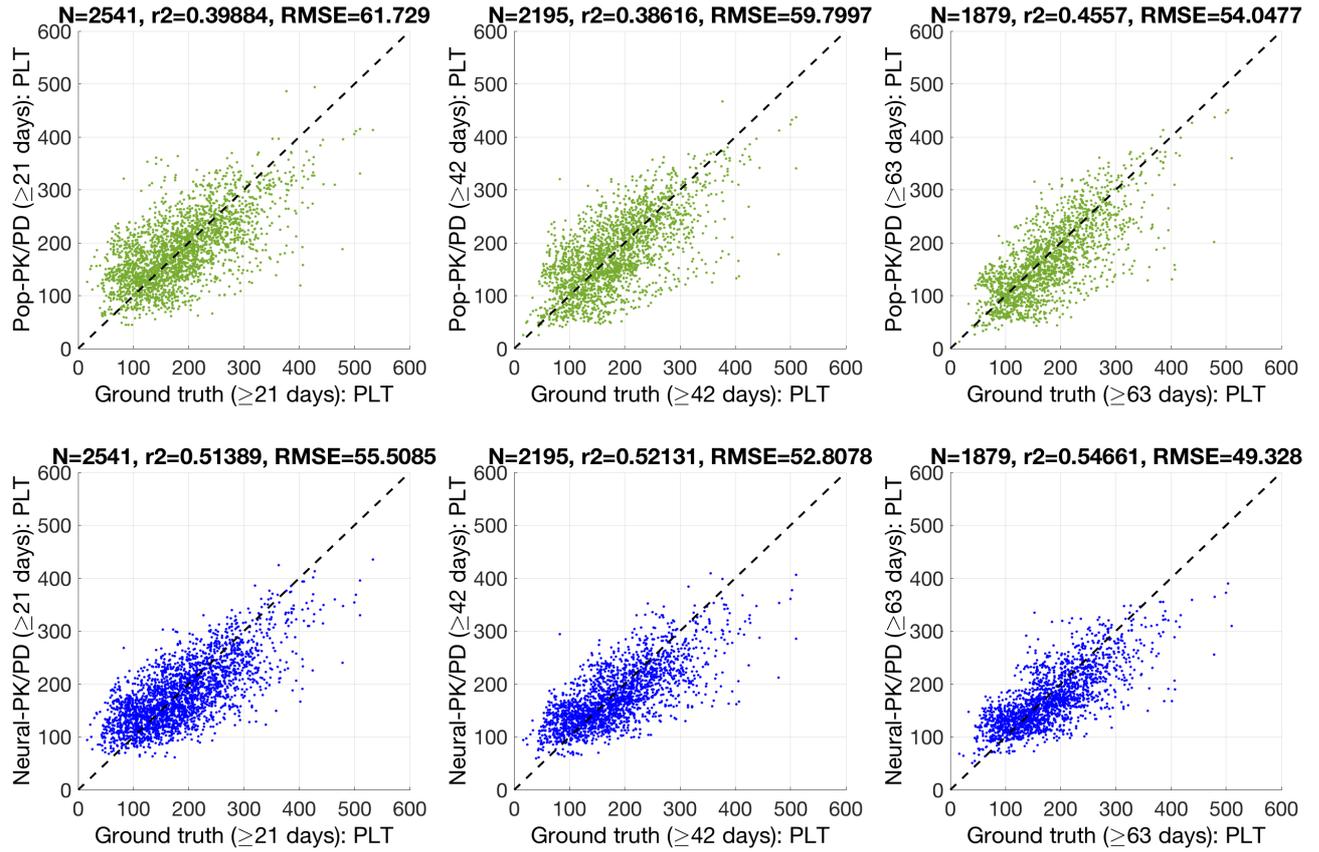

*Figure 4. Comparison of the model prediction versus ground truth, using the pop-PK/PD (top row) and neural-PK/PD models (bottom row). We consider the scenarios of observing platelet data within cycles 1 to 3 of treatment (i.e., corresponding to time < 21, 42 and 63 days respectively) and attempting to predict the future platelet time course (i.e., corresponding to t ≥ 21, 42 and 63 days respectively). In each case, N refers to the number of predictions made for the corresponding scenario. The prediction results show that neural-PK/PD model has a numerically higher r2 and lower RMSE in predicting future platelet counts than the pop-PK/PD model.*

| Case | Observation | | Prediction | | Pop-PK/PD | | Neural-PK/PD | |
|---|---|---|---|---|---|---|---|---|
| | Window (day) | # of obs. | Window (day) | # of pred. | *r2* | *RMSE* | *r2* | *RMSE* |
| (a) | 0 < t < 21 | 413 | t ≥ 21 | 2541 | 0.40 | 61.73 | **0.51** | **55.51** |
| (b) | 0 < t < 42 | 759 | t ≥ 42 | 2195 | *0.39* | *59.80* | **0.52** | **52.81** |
| (c) | 0 < t < 63 | 1075 | t ≥ 63 | 1879 | 0.46 | 54.05 | **0.55** | **49.33** |
| | | | | | | | | |
| (d) | 0 < t < 21 | 413 | t ≥ 42 | 2195 | - | - | *0.45* | *56.36* |

*Table 1. Comparison of predictive performance for pop-PK/PD versus neural-PK/PD. Cases (a) to (c) compare pop-PK/PD with neural-PK/PD based on identical sets of observation and prediction data, the results demonstrate a superior performance for the latter methodology. In case (d), neural-PK/PD predictions using 413 early observations (from 0 < t < 21 days) is shown to give rise to numerically higher performance as compared to pop-PK/PD (case (b), italicized) which utilizes nearly doubled number of observations (759) from the observation window 0 < t < 42 days.*

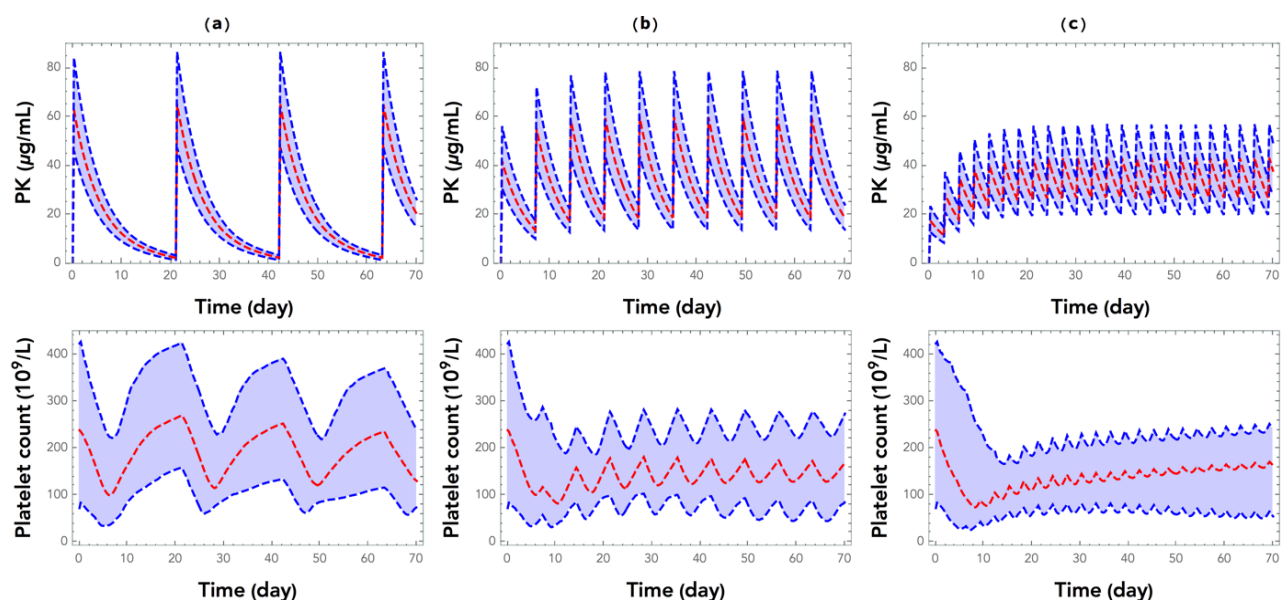

*Figure 5. Neural-PK/PD performing population simulations using the (n=532) training patients: (a) Q3W dosing at 3.6 mg/kg of T-DM1; (b) Q1W dosing at 2.4 mg/kg of T-DM1; (c) Q3D dosing at 1.0 mg/kg of T-DM1. The dashed red lines show the population medians, with the dashed blue lines showing the 5th to 95th percentiles of the training patients. Despite not having any patient given Q3D dosing in the training set, the model was able to generate realistic profiles that demonstrate the expected PK accumulation and frequent platelet fluctuations as a result of the proposed model architecture which is based upon PK/PD principles.*

**Discussion**

In this work, we present a novel application of deep learning approach to predict drug concentration and response time course, for the purpose of enabling personalized predictions in individual patients and predict the effects of different dose schedules. Our work represents the first effort for a head-to-head comparison of a pop-PK/PD model with a deep learning model that we are aware of. Being distinct methodologies analyzing drug concentration/response data and making clinical inferences, the traditional pop-PK/PD and the proposed novel neural-PK/PD approaches have their respective pros and cons. While the former approach incorporates basic pharmacological principles that enable meaningful interpretation, model development can be time consuming due to the need for human-driven testing of alternative models and various diagnostic evaluations, thus hampering the ability to use it for real-time predictions or complex data sets. On the other hand, while neural-PK/PD is driven by data and can be generated algorithmically with little direct human supervision thus saving time and human resources, it remains to be validated on many datasets, and we hope this study will spur such analyses in the near future. In addition, the incorporation of variability into neural-PK/PD models so as to enable clinical trial simulations remain to be further developed.

Although deep learning models have been shown to have impressive ability to directly learn from vast amounts of data and enable predictions with little human intervention, existing techniques are known to have certain drawbacks [25, 26, 27]. Firstly, many deep learning models tend to be data hungry [28] and their applicability to data sets of moderate size that arise from clinical trial settings may be limited. Secondly, it is known that deep learning models may use features in the training data that do not generalize well to other situations [26], thereby potentially lowering the utility of such models especially for prediction of untested scenarios. We believe that the incorporation of key domain specific concepts into deep learning frameworks is an effective way to combine human and machine intelligence and can be crucial for the successful applications of such models. Within the domain of longitudinal analysis of patient treatment data, we propose the incorporation of well-established PK/PD principles into neural-ODE framework [7] by an appropriate choice of network architecture as a way to ensure that these models would serve well for the purpose of individualized predictions of the effects of untested dosing regimen. We demonstrate the feasibility of such novel approach for drug concentration and response prediction based on a legacy clinical trial data set consisting of more than 600 patients. While the results shown in this work demonstrate the proposed neural-PK/PD modeling approach can improve upon the current "gold-standard" pop-PK/PD modeling on prediction metrics, the clinical implications and meaningfulness of such improvements need to be further assessed in future work. The importance of using machine to aid human to build models will increase as technological advancements lead to an ever-increasing set of diagnostic modalities and monitoring devices that generate growing amounts of patient treatment data in real time; relying on human insight alone to generate models that describe the observed data will become increasingly challenging in the digital age. With the joint requirements of automating modeling to save time and human resources, while ensuring the pharmacological meaningfulness and the predictability of the generated models to inform dosing considerations, it is important to bridge modern deep learning and traditional PK/PD methodologies. Although machine learning is currently not part of clinical pharmacology modeling applications, it is appreciated that a number of opportunities exist for leveraging both paradigms [29]. We propose that the "artificial intelligence (AI) enabled clinical pharmacologists" of the future [30] would tap into neural-PK/PD as one of the advanced analytics tools to understand and predict drug concentration and response for dosing recommendation. Additionally, as neural-PK/PD models "speak" the language of neural networks, it could be combined with other deep learning models for novel data types and can as well be integrated into digital devices for automated longitudinal patient monitoring to generate treatment insights.

While the proposed neural-PK/PD formulation demonstrates encouraging results, there are several areas that remain topics for future work. The current model in the form of a recurrent architecture that takes fixed time steps in the numerical approximation of the underlying ODE system. Various efforts are underway to generalize the current framework by incorporating neural ODE solvers that take adaptive step sizes (e.g., [7]), so as to account for the arbitrary time values that may be present in the data set (i.e., measurement and dosing times). While the data set shown in this work involves time points that take on integer values (in the unit of day) and hence directly applicable to the proposed model, the incorporation of adaptive step size is expected to provide increased efficiency and accuracy in the general setting. For this extension, although the "ODE" sub-module shown in Fig. 1 would need to be replaced by an adaptive ODE solver [7], the architecture of the proposed neural-PK/PD model would remain the same. Finally, in the current work we did not use any baseline features of the patients (such as demographics, pre-treatment lab values, etc) for a head-to-head comparison with the original pop-PK/PD model [17] which similarly did not include any covariate effects. Incorporation of the available baseline features may better characterize patient heterogeneity and improve the model predictivity, remains a topic for future work. In general, this novel neural-PKPD approach also need to be tested with more drugs, indications, as well as various type of therapeutic and adverse responses, to further support the robustness and predictivity of the approach in clinical settings. To conclude, the promising results shown in this work and the potential advantages offered by this novel neural-PK/PD methodology warrants further development and testing. We expect that the development and deployment of neural-PK/PD models would help to facilitate clinical drug development and advance personalized medicine in clinical practice.

**Methods**

In this work, the platelet count measured in patients' blood corresponds to the PD variable whose time course we attempt to describe in relation to the PK (or drug concentration). We compare the neural-PK/PD model with pop-PK/PD in terms of their ability to predict future platelet counts from early data. We do so by taking 80% of the total patients in the data as the training set and the remaining 20% as the test set, in a manner following the machine learning paradigm [31, 32]. In particular, we consider a hypothetical scenario whereby we have obtained data from a prior clinical study (i.e., training set) and built a model on it; subsequently, we would like to apply the model to a new trial of a similar patient population and predict the individual patient future platelet time course from having observed only the early data. While pop-PK/PD models are typically evaluated based on their ability to describe the available data with parsimony considerations [4, 5] rather than for their ability to make temporal prediction, the current approach is one way to compare the models on an equal footing.

*Data*

The data consists of longitudinal platelet response from 665 patients receiving T-DM1, including patients from one Phase I study (TDM3569g), three Phase II studies (TDM4258g, TDM4374g and TDM4450g) and one Phase III study (TDM4370g). Please refer to [18, 17, 21] for further details regarding the clinical studies and the patient demographics. All patients provided written informed consent. Patients in the Phase II/III studies all received T-DM1 at 3.6 mg/kg Q3W. Patients in the Phase I study received various doses and schedules of T-DM1: 0.3 mg/kg Q3W (n=3), 0.6 mg/kg Q3W (n=1), 1.2 mg/kg Q3W (n=1), 3.6 mg/kg Q3W (n=14), and 4.8 mg/kg Q3W (n=3); 1.2 mg/kg weekly (Q1W) (n=3), 1.6 mg/kg Q1W (n=2), 2.0 mg/kg Q1W (n=1), 2.4 mg/kg Q1W (n=13), and 2.9 mg/kg Q1W (n=2).

*Data processing for neural network*

From the entire data set, an 80%-20% split of patients was made to divide them into the training (n=532) and testing (n=133) subsets. Data normalization was performed by taking the complete time course from the training set, and the normalization factors are computed so as to ensure each variable (i.e., data column) is normalized to have mean of 0 and standard deviation of 1; subsequently, the same scaling factors are applied to transform the testing set.

To combat overfitting, data augmentation on the set of training patients was performed as follows: we take the union of the following data sets, where in each case the notation "input" → "output" denotes that the network is fed "input" into the network port "PKPDData" and asked to predict the "output" as the prediction target:

- Complete time course: for training patient *i*,
  $\{PK^i(t), Dosing^i(t), PD^i(t)\}_{0 \leq t < \infty} \rightarrow \{PK^i(t), PD^i(t)\}_{0 \leq t < \infty}$
- Observation data up to day 21: for training patient *i*,
  $\{PK^i(t), PD^i(t)\}_{0 \leq t < 21}, \{Dosing^i(t)\}_{0 \leq t < \infty} \rightarrow \{PK^i(t), PD^i(t)\}_{0 \leq t < \infty}$
- Observation data up to day 35: for training patient *i*,
  $\{PK^i(t), PD^i(t)\}_{0 \leq t < 35}, \{Dosing^i(t)\}_{0 \leq t < \infty} \rightarrow \{PK^i(t), PD^i(t)\}_{0 \leq t < \infty}$
- Observation data up to day 42: for training patient *i*,
  $\{PK^i(t), PD^i(t)\}_{0 \leq t < 42}, \{Dosing^i(t)\}_{0 \leq t < \infty} \rightarrow \{PK^i(t), PD^i(t)\}_{0 \leq t < \infty}$

- Observation data up to day 63: for training patient *i*,

$$\{PK^i(t), PD^i(t)\}_{0 \leq t < 63}, \{Dosing^i(t)\}_{0 \leq t < \infty} \rightarrow \{PK^i(t), PD^i(t)\}_{0 \leq t < \infty}$$

After performing the above described data augmentation by cutting the data at different observation times, we have 532×5=2660 set of augmented patient records. The aim of the data augmentation process is to enrich the training data set so as to force the neural network to achieve to goal of enable predictions in future time based having only the early observation data as well as the full dosing record.

*Neural-PK model and training*

The neural-PK model was constructed to reproduce the T-DM1 concentration time-course predicted by the pop-PK model [18]. The schematic diagram of the model is shown in Supplementary Fig. 1(a). The input data enters the network through the port "PKPDData"; the data consists of a variable number of rows (denoted by $n_1$ in Supplementary Fig. 1(a)) and 5 columns, which are: (1) time-after-dose, (2) time, (3) PK, (4) platelet count, and (5) amount dosed. In particular, for our neural-PK model, the platelet count is not used under the assumption that PK is not influenced by PD. Hence, only the columns 1, 2, 3 and 5 are selected via the "PKPart" network. The selected data columns are then fed into the "PKEncoder", which at its core contains a Gated Recurrent Unit (GRU) [33] with a state size of 20 which finally outputs a 4-dimensional vector after having gone through the sequence of data rows; for further details please refer to Supplementary File 1.

After having encoded the PK data, the model simulation is performed via the "PKODE" sub-module, which is a neural network of recurrent architecture that performs the forward Euler time steps [34, 7]. Further details of this recurrent architecture are shown in Supplementary Fig. 1(b). As there is no drug concentration in the patients prior to dosing, the network port "InitState" (as shown in Supplementary Fig. 1(a)) is set to a vector of zeros. Dosing enters explicitly in this recurrent network via the network port "Dose". Details on how "Dose" enters into the model is further illustrated in Supplementary Fig. 1(b): the dosed component of the PK state is incremented by the given dosages provided in the data set; note that the non-dosed component is not varied via the padding layer [23] "pad 0". The network "PKVF" aims to approximate the vector field of the PK model (see Fig. 1(c)). The layer "$\Delta t*$" performs element-wise multiplication by the step size (1/4 day in the current implementation) corresponding to the discretized time rate of change of the PK system. Finally, the rectified linear unit [23] "ReLU" ensures that the PK state remains non-negative. The model construction was implemented in Wolfram Mathematica [35] and is provided in the Supplementary File 1.

Through backpropagation algorithm [23], the trainable weights of "PKEncoder" and "PKODE" sub-modules are iteratively refined to minimize the $L_2$ loss function between the observed data and model outputs. As previously mentioned, we used an 80-20 split of the total patient data. Data augmentation as described in the previous section was performed. Note that once selected, the test patients are never used in the training phase but kept aside for the final evaluation of the network. The ADAM optimizer [23] was used to train the neural network for a total of 2000 epochs. The trained network is then evaluated on the n = 133 test patients, and comparison to the ground truth showed good performance, with r2 = 0.98, correlation coefficient of 0.99 and RMSE=2.67.

*Pop-PK/PD model and prediction*

We used a pop-PK/PD model with structure as described in [17, 18]. In particular, the model consists of a two-compartmental PK model (with 2 state variables and 4 parameters) and the platelet dynamics is described using 6 state variables and 10 parameters that are variable between patients (i.e., where inter-individual variability is allowed) [5]. The model was built sequentially [20, 5], with PK parameters first estimated and subsequently followed by the PD parameters using the First Order Conditional Estimation (FOCE) method in NONMEM 7.3.0 [36].

For the pop-PK/PD approach, we process data in the following manner: (a) for patients in the training set, we leave the whole observation and dosing data intact; (b) for patients in the test set, we keep all the dosing data but only retain PK & PD observation data within the initial window t < tObs (i.e., setting "DV" to empty and "MDV"=1 in the NONMEM [36, 5] data set). From this input data, NONMEM [36] was used to carry out estimation of the model parameters both at the population level and the individual level. From the result of the estimation, NONMEM produces amongst its outputs the individual predictions ("IPRED") for platelet dynamics. We make predictions for the unseen portion (i.e., t ≥ tObs) for patients in the test set and compare them to the data.

*Neural-PD network and training*

Having first trained the neural-PK net, we subsequently copied over the trained weights from "PKEncoder" and "VFnet" of Supplementary Fig. 1 onto the corresponding PK components of the neural-PK/PD network shown in Fig. 1. These weights are then frozen from any further training. The $n_1 \times 5$ input data that enters the network port "PKPDData" is then selected for the following 4 columns via the "PDData" sub-module shown in Fig. 1: (1) time-after-dose, (2) time, (3) PK and (4) platelet count; that is, the dosing column is dropped. The implicit assumption being made is that it is the drug concentration which drives the pharmacodynamic effect, but dosing does not directly mediate the pharmacodynamic effect; this is a standard assumption made in PK/PD modeling. The selected data columns are then fed into "PDEncoder", which at its core contains a Gated Recurrent Unit (GRU) [33] with a state size of 50 and outputs a 10-dimensional vector. Finally, both the outputs of "PKEncoder" (4-dimensional vector) and "PDEncoder" (10-dimensional vector) are fed into the "ODE" sub-module, which simulates both the PK and PD components of the model using those encoded vectors, driven by the dosing coming in via the "Dose" network port shown in Fig. 1. In our model, we represent the PD (platelet) dynamics using a 4-dimensional ODE state vector. The underlying assumption that PK drives PD (but not vice versa) is encoded into the network architecture of Fig. 1(b). For the PD dynamics, the corresponding 4-dimensional ODE state vector needs to be initialized prior to the first dose being given; this is in contrast to the PK part of the model, where prior to dosing the drug concentrations are identically zero. The combined network (Fig. 1) takes as an input the initial platelet state obtained from looking at the first platelet count after the start of treatment ($t \geq 0$); this data pre-processing was done for all patients and enters through the network port "InitState". However, simply taking the platelet estimate prior to treatment as the first observed platelet data is inaccurate as some patients had a marked platelet drop right after the start of treatment; hence we use the "ICNet" to help improve the estimation the initial platelet count. This was done by estimating a scalar from the early observed data and then replicating it to the 4-dimensional platelet state vector. The need to estimate the initial platelet value from the time course data is similar to that of the pop-PK/PD mylosuppression models [19, 17]. The "ICNet" consists of a bi-directional [37] GRU with a state size of 10; see Supplementary File 1 for further details of the implementation. The network port "ObsState" outputs both the PK and PD (platelet count) predictions at each time step. In the sequential training approach, only the PD component is compared to the platelet data, based on the $L_2$ loss function (which computes the mean squared error). The ADAM optimizer [23] was used to train the neural-PK/PD network for a total of 3000 epochs until the training loss no longer improves.

*Model Prediction Benchmarking*

We ensure that both the pop-PK/PD and neural-PK/PD are compared against the ground-truth platelet data in an identical manner on the test patients, with respect to both r2 and RMSE.

*Simulating dosing regimen*

While most patients in the training set were given Q3W dosing, a few were given doses at Q1W frequency. Nevertheless, we can use the neural-PK/PD model to simulate the scenario whereby all the patients were given T-DM1 at the dose 3.6 mg/kg Q3W as follows. Firstly, the complete training patient data were fed into the network port "PKPDData" to obtain the encoded parameters outputs from network sub-modules "PKEncoder" and "PDEncoder". Then, we feed the desired regular dosing of 3.6 mg/kg Q3W into the "Dose" network port. The $5^{th}$, $50^{th}$ and $95^{th}$ quantiles of the neural-PK/PD simulated results from the training patients were then computed to generate the plots shown in Fig. 5(a). A similar procedure was used to simulate T-DM1 at the desired dosing regimens in the training patients to generate Fig. 5(b) and 5(c).

**Acknowledgements**

The authors would like to thank Dan Lu, Logan Brooks, Gengbo Liu, Kai Liu, Kaiwen Deng and Amita Joshi for their discussions and helping to make this work possible.